\PassOptionsToPackage{table,xcdraw,dvipsnames}{xcolor}
\documentclass{article}

\PassOptionsToPackage{sort, numbers, compress}{natbib}

\usepackage[preprint]{neurips_2025}

\usepackage[utf8]{inputenc} 
\usepackage[T1]{fontenc}    
\usepackage[colorlinks,citecolor=blue]{hyperref}       
\usepackage{url}            
\usepackage{lipsum}         

\usepackage{amsfonts}       
\usepackage{amssymb}
\usepackage{mathtools}
\usepackage{amsthm}
\usepackage{bm,verbatim}
\usepackage{bbm}
\usepackage{physics}

\usepackage{nicefrac}       
\usepackage{microtype}      
\usepackage[most]{tcolorbox}

\usepackage[capitalize,noabbrev]{cleveref}

\usepackage[shortlabels,inline]{enumitem}  

\usepackage{relsize}
\usepackage{graphicx}       
\usepackage{booktabs}       
\usepackage{wrapfig}        
\usepackage{multirow}       
\usepackage{subcaption}
\usepackage{ctable}         
\usepackage[normalem]{ulem} 
\usepackage{colortbl}
\usepackage{tablefootnote}
\usepackage{algorithmic,algorithm}

\usepackage[T1]{fontenc}

\usepackage[utf8]{inputenc}

\usepackage{microtype}

\usepackage{inconsolata}

\usepackage{graphicx}

\usepackage{tcolorbox}
\tcbuselibrary{breakable}
\usepackage{booktabs}
\usepackage{graphicx}
\usepackage{multirow}
\usepackage{amsmath}
\usepackage{float}
\usepackage{makecell}
\usepackage{enumitem}
\usepackage{colortbl}

\newtheorem{proposition}{Proposition}
\newtheorem{corollary}{Corollary}

\begin{document}
\title{Multimodal Reasoning Agent for Zero-Shot Composed Image Retrieval}

%

\author{
  \textbf{Rong-Cheng Tu\textsuperscript{1}},
  \textbf{Wenhao Sun\textsuperscript{1}},
  \textbf{Hanzhe You\textsuperscript{2}},
  \textbf{Yingjie Wang\textsuperscript{1}},
\\
  \textbf{Jiaxing Huang\textsuperscript{1}},
  \textbf{Li Shen\textsuperscript{3}},
  \textbf{Dacheng Tao\textsuperscript{1}\footnotemark[1]}
\\
  \textsuperscript{1}College of Computing and Data Science, \\ Nanyang Technological University, Singapore, Singapore
\\
  \textsuperscript{2}School of Information Science and Technology, \\ University of Science and Technology of China, Hefei, China
\\
  \textsuperscript{3}Sun Yat-sen University Shenzhen Campus, \\ School of Cyber Science and Technology, Shenzhen, China
\\
    \texttt{rongcheng.tu@gmail.com,} \texttt{dacheng.tao@ntu.edu.sg}
}

\renewcommand{\thefootnote}{\fnsymbol{footnote}}

\footnotetext[1]{\quad Co-corresponding authors}

\maketitle

\begin{abstract}
Zero-Shot Composed Image Retrieval (ZS-CIR) aims to retrieve target images given a compositional query—consisting of a reference image and a modifying text—without relying on annotated training data. 
Existing approaches often generate a synthetic target text using large language models (LLMs) to serve as an intermediate anchor between the compositional query and the target image. Models are then trained to align the compositional query with the generated text, and separately align images with their corresponding texts using contrastive learning.
However, this reliance on intermediate text introduces error propagation, as inaccuracies in query-to-text and text-to-image mappings accumulate, ultimately degrading retrieval performance.
To address these problems, we propose a novel framework by employing a Multimodal Reasoning Agent (MRA) for ZS-CIR. MRA eliminates the dependence on textual intermediaries by directly constructing triplets, <reference image, modification text, target image>, using only unlabeled image data. By training on these synthetic triplets, our model learns to capture the relationships between compositional queries and candidate images directly.
Extensive experiments on three standard CIR benchmarks demonstrate the effectiveness of our approach. On the FashionIQ dataset, our method improves Average R@10 by at least 7.5\% over existing baselines; on CIRR, it boosts R@1 by 9.6\%; and on CIRCO, it increases mAP@5 by 9.5\%. 
\end{abstract}

\section{Introduction}
Traditional image retrieval approaches, whether content-based \cite{dcgmh,stbmh} or text-based \cite{daphcm,rcitr}, often struggle to handle complex user queries involving visual and textual elements. Composed Image Retrieval (CIR) \cite{tirg, wen2024simple, li2023dual,xu2023multi, bai2023sentence} addresses this limitation by allowing users to query using an example image together with a natural language modification. This combined query allows fine-grained control over retrieval results: the reference image anchors the query in a concrete visual example, while the text specifies how to transform or refine it to match the desired target.
Despite their demonstrated effectiveness, conventional CIR methods~\cite{TianYuxin23Image,cirr,circlt,bai2023sentence} heavily depend on manually annotated training triplets comprising a reference image, modifying text, and a target image. This dependence severely constrains their scalability, as generating high-quality labeled triplets is expensive and labor-intensive, making adaptation to new domains challenging.

To address this limitation, Zero-Shot CIR (ZS-CIR) methods~\cite{fti4cir,pic2word,mllmi2w,mcl} have emerged, aiming to eliminate reliance on explicit annotation. Early approaches in ZS-CIR methods typically train lightweight adapters for frozen vision-language models (VLMs) \cite{clip,li2023blip,scl_vl,glscl} to convert visual features of reference images into pseudo-text embeddings. This conversion simplifies compositional queries (reference image + modifying text) into unified textual representations, which can then be processed directly by pre-trained VLMs for cross-modal retrieval.
More recent advances \cite{tansagg,mcl,instructcir} incorporate large language models (LLMs) and multi-modal LLMs (MLLMs)~\cite{llama,llma_survey,llava,spagent} to further enhance ZS-CIR performance. These approaches generate synthetic training triplets consisting of a reference image, modification text, and target text. Leveraging these synthetic triplets, the models explicitly learn to align compositional queries with corresponding textual descriptions, while simultaneously aligning images with their matched captions using contrastive learning. By bridging compositional queries and candidate images through intermediate textual embeddings, these methods effectively map them into a unified semantic space, enabling composed image retrieval.

However, these recent ZS-CIR methods \cite{fromage,mcl} are particularly susceptible to error propagation, significantly hindering their retrieval performance. Specifically, the generated target texts often fail to fully reflect the intricate semantic nuances of the compositional queries, leading to incorrect compositional representations. Additionally, the training image-text pairs typically originate either from automatic generation via MLLMs \cite{gpt,llava,minicpm} or from noisy web sources with captions that are frequently generic, ambiguous, or inadequately descriptive. Such noisy textual supervision introduces cumulative errors during both compositional query-to-text alignment and subsequent text-to-image mapping stages. Consequently, these cumulative inaccuracies severely impair the model's capability in accurately capturing the relationship between compositional queries and target images.

Given these limitations, a fundamental question arises: 
Can we bypass the intermediate textual representation and directly align compositional queries with target images, by constructing high-quality <reference image, modifying text, target image> triplets directly from unlabeled image data without manual annotations?
Achieving this requires solving two key challenges: (1) The reference image and target image should share appropriate semantic similarity. If the images are too similar, the modifying text becomes redundant, reducing CIR to image-to-image retrieval. Conversely, if the images are too dissimilar, the modifying text functions as a target image caption, misguiding the model into treating CIR as a text-to-image retrieval task. Both scenarios result in biased representations and harm retrieval performance. (2) The combination of the reference image and modification text should precisely describe the target image’s content, ensuring that retrieval models learn accurate compositional representations.

To address these challenges, we propose a novel Multimodal Reasoning Agent-based CIR (MRA-CIR) framework that constructs high-quality triplets directly from unlabeled image data, enabling fine-tuning of VLMs for ZS-CIR tasks. Specifically, to ensure appropriate semantic similarity between reference and target images, we first leverage a pre-trained VLM to extract image embeddings and compute pairwise similarities. Instead of selecting the most similar images, which would trivialize CIR into standard image retrieval, we identify moderately similar images, ensuring that meaningful but non-trivial transformations are required. Next, we propose a context-aware semantic reasoning strategy that employs a Multimodal Reasoning Agent (MRA)—an MLLM MiniCPM-VL-2\_6 \cite{minicpm} equipped with advanced capabilities in semantic understanding and visual comparison—to generate accurate modification texts. The MRA identifies key differences between the reference and target images and then formulates precise textual descriptions that describe how the reference image can be transformed into the target. These outputs are then used to construct high-quality triplets (<reference image, modifying text, target image>) that are highly aligned with the objectives of the CIR task. 
By fine-tuning the adopted VLM with an InfoNCE loss computed via token-level maximum cosine similarity over these high-quality triplets, our approach explicitly captures the compositional alignment between queries and candidate images, effectively guiding the model to associate query features with their correct targets. Moreover, we provide a rigorous theoretical analysis showing that our loss function optimizes a valid lower bound of the standard InfoNCE loss \cite{infonce}, thereby offering principled justification for the effectiveness of our learning strategy.
Extensive experiments on three benchmark datasets demonstrate that MRA-CIR significantly outperforms existing state-of-the-art ZS-CIR methods, achieving superior retrieval accuracy and robustness.

\section{Related Work}
Standard image retrieval techniques typically operate in unimodal settings—either identifying images based on visual resemblance \cite{uhscm,stbmh,wglhh,psldh}, or retrieving results that align with a standalone textual description \cite{ucmhmi,dcph,uchstm,rcitr}. However, such approaches are often inadequate for tasks where user intent is inherently multimodal, involving both a visual reference and a desired transformation. {Composed Image Retrieval (CIR)} \cite{TianYuxin23Image,searle,cirr,pic2word,keds,fti4cir,tirg} has emerged to address this challenge by enabling queries that combine an image with a free-form textual modifier. This setup allows users to convey not just what they are looking for, but how it should differ from an example—supporting nuanced, fine-grained control in open-domain retrieval scenarios.
\subsection{Composed Image Retrieval (CIR)}
Composed Image Retrieval (CIR)~\cite{wen2024simple, zhang2023enhance, TianYuxin23Image,cirr,circlt,bai2023sentence,tirg,vdg,wen2024simple,tgcir,limn} aims to learn a joint representation that fuses a reference image and a relative textual description in order to retrieve the target image. For example~\cite{tirg}, introduced a residual gating mechanism to combine reference image features with modifying text. VDG \cite{vdg} utilizes labeled triplets to train an MLLM, which subsequently generates additional triplets alongside the labeled ones to further enhance the training of the VLM for the CIR task. 
TG-CIR \cite{tgcir} exploits a knowledge distillation mechanism to guide conflict modeling in multimodal queries through target image integration, while also refining the metric learning process.
LIMN+ \cite{limn} introduces a self-training framework that iteratively generates high-quality triplet samples, thereby alleviating data scarcity and enhancing generalization capabilities
Although these methods have demonstrated promising performance, they typically depend on extensive collections of labeled triplets <reference image, text, target image>. However, obtaining such annotations is both labor-intensive and expensive, which hinders the scalability of CIR systems across new domains and applications.
\subsection{Zero-Shot Composed Image Retrieval}
\label{sec:zscir}
Zero-Shot Composed Image Retrieval (ZS-CIR) methods \cite{pic2word,searle,keds,fti4cir,mcl} aim to bypass the reliance on manually annotated triplets by learning how to integrate visual and textual information from unlabeled or minimally labeled data. 
An influential paradigm in ZS-CIR is to map the reference image into one or more pseudo-text tokens, then concatenate these tokens with the user-provided text query to perform cross-modal retrieval. Early pioneer works~\cite{pic2word,searle} propose training a visual adapter on a frozen Vision-Language Model (VLM), transforming image embeddings into pseudo-word embeddings. 
KEDs \cite{keds} implicitly capture reference image attributes by leveraging a database that enriches pseudo-word tokens with relevant images and captions, highlighting shared attribute information across different aspects.
Contex-I2W~\cite{contex12w} builds upon this idea by introducing an image-to-word mapping network that leverages manipulation descriptions and learnable queries for context-aware visual filtering.

Inspired by the remarkable semantic understanding and instruction-following capabilities of Large Language Models (LLMs), several recent approaches integrate LLMs to enhance ZS-CIR. For instance, MLLM-I2W~\cite{mllmi2w} employs a multimodal LLM to select subject words and enrich textual descriptions, thus translating the reference image into more expressive pseudo-text tokens. Other methods, such as LaSCo \cite{case} or TransAgg~\cite{tansagg}, propose using GPT-3~\cite{gpt} or related models \cite{llama,llava} to construct synthetic CIR triplets directly from existing QA or caption datasets. MCL~\cite{mcl} also generates triplets <\text{reference image}, \text{text condition}, \text{target caption}> via a multimodal LLM, which are then employed to fine-tune a model for compositional retrieval.

While these methods have demonstrated noteworthy progress, they often rely on intermediate representations or additional modules (e.g., pseudo-text tokens, generated target text) to bridge the gap between composed query and target image. Such multi-step conversions can introduce cumulative errors that degrade retrieval accuracy. 
In contrast, we propose leveraging a Multimodal Reasoning Agent (MRA) to automatically construct triplets <\text{reference image}, \text{modification text}, \text{target image}> from unlabeled images. This direct approach mitigates the risk of error propagation by avoiding multiple conversion steps.

\section{Proposed Method\label{sec:task_formulation}}
\begin{figure*}[]
    \centering
    \includegraphics[width=\linewidth]{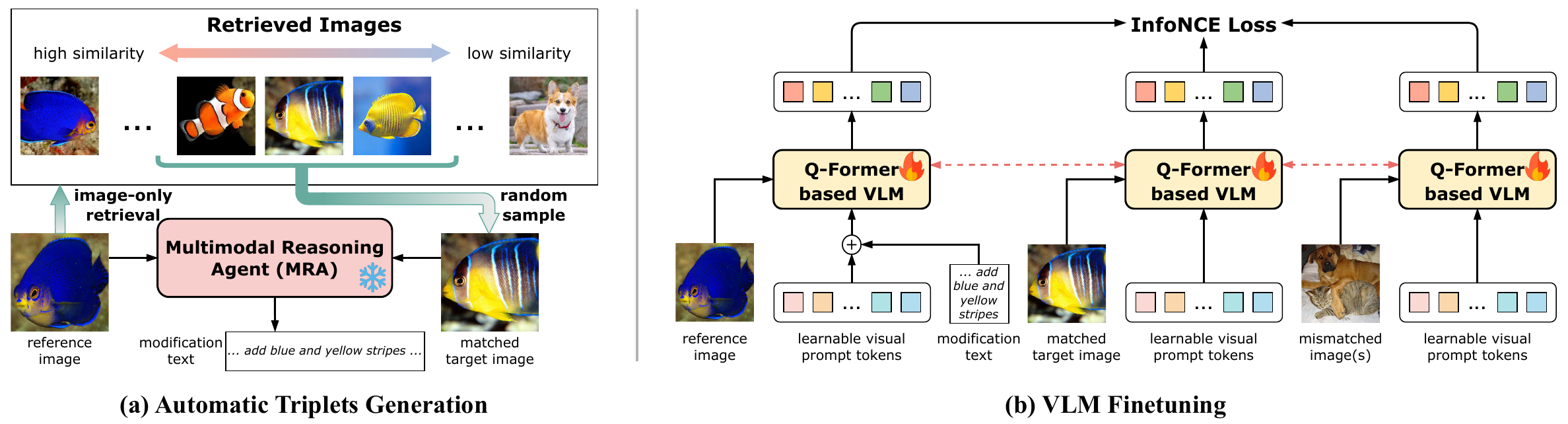}
    \caption{The illustration of (a) automatic triplets generation and (b) the framework of MRA-CIR.}
    \label{fig:framework}
\end{figure*}

This section delineates the proposed MRA-CIR framework where an overview is shown in Figure \ref{fig:framework}. Section \ref{section:data_generation} introduces how to generate the training triplets through the MRA. In Section \ref{section:finetune_vlm}, we introduce how to fine-tune VLM for the CIR task.

\subsection{Automatic Triplets Curation}
\label{section:data_generation}
As shown in Figure \ref{fig:framework} (a), to construct high-quality triplets <reference image, modifying text, target image>, we first construct the reference--target image pairs with moderate similarities and then generate a modification text for each data pair through the multimodal reasoning agent (MRA).
\paragraph{Moderate Similarity-Based Target Image Selection.}
Given an unlabeled image dataset $\{x_i\}_{i=1}^n$, where $n$ denotes the total number of images, we treat every image in the unlabeled dataset $\{x_i\}_{i=1}^n$ as a potential reference image, and need to pair it with a target image. To achieve this goal, we first extract token-level feature representations for each image $x_i$ using a pre-trained BLIP-2 model \cite{li2023blip}. Let $\{\boldsymbol{f}_i^k\}_{k=1}^m$ be the token-level feature vectors of the $i$-th image $x_i$, where $\boldsymbol{f}_i^k$ is the $k$-th token and $m$ is the total number of tokens. To measure the similarity between two images, we compute the maximum cosine similarity over all token pairs and then average across tokens:
\begin{equation}
\begin{aligned}
h_{ij} &= \frac{1}{m}\sum_{k=1}^m \max_{1 \le l \le m} \frac{(\boldsymbol{f}_{i}^k)^T \cdot \boldsymbol{f}_{j}^l}{\|\boldsymbol{f}_{i}^k\|_2 \|\boldsymbol{f}_{j}^l\|_2},
\end{aligned}
\label{eq:token_sim}
\end{equation}
where $\|\cdot\|_2$ denotes the L2 norm of a vector, and $h_{ij}$ is the token-level-based similarity across between images $x_i$ and $x_j$. This token-level approach can capture both local and global similarities, offering a more fine-grained measure of semantic relatedness.

Once the pairwise similarities are computed, we rank the remaining images for each image \(x_i\) which is used as a reference image. We then randomly pick an image from the subset of similar images ranked between \(q_1\) and \(q_2\) as the target image \(y_i\), where \(q_1\) and \(q_2\) are hyper-parameters satisfying \(1 < q_1 < q_2 \le n\). This design aims to strike a balance between two key factors: 1) \textit{Avoid trivial modifications.} Selecting the most similar image as the target might only involve minor changes (e.g., slight variations in color or texture), leading the model to treat the compositional retrieval task as if it were image-to-image matching. 2) \textit{Avoid unrelated images.} Selecting images with extremely low similarity would make the modifying text a direct description of the target, thus reducing the problem of text-to-image retrieval.

In both extreme cases, the model tends to rely disproportionately on either the visual modality or the textual modality when extracting composed query features, leading to biased representations and degrading retrieval performance. 
By selecting target images with a moderate level of similarity, each constructed triplet compels the model to integrate information from both the reference image and its textual modifications to accurately capture the target image’s semantic content. As a result, the model more effectively fuses useful features from both modalities, producing higher-quality composite representations and ultimately improving retrieval performance.

\paragraph{Modifying Text Generation via MRA.}

Having identified a moderately similar target image $y_i$, our goal is to generate a concise modifying text $t_i$ describing how the reference image $x_i$ can be transformed into $y_i$. 
We use the MRA to generate the modification text.
It is based on an MLLM \cite{internvl} that interprets the semantic content of each image and identifies their differences.

A straightforward approach is to input $\bigl(x_i, y_i\bigr)$ directly into the MRA with a prompt $P_m'$ for the modification text: $t_i = \mathrm{MRA}\bigl(x_i, y_i, P_m'\bigr).$
While this intuitive approach is simple, it often overlooks subtle context or fine-grained differences between $x_i$ and $y_i$. 
To tackle this issue, we employ a two-step strategy to generate precise, context-aware modification text.

First, we prompt the MRA to generate captions $c_{x_i}$ and $c_{y_i}$ for $x_i$ and $y_i$, respectively:
\begin{equation}
c_{x_i} = \mathrm{MRA}\bigl(x_i, P_c\bigr), 
\quad 
c_{y_i} = \mathrm{MRA}\bigl(y_i, P_c\bigr),
\label{eq:caption_generation}
\end{equation}
where $P_c$ is the prompt for guiding the MRA to describe each image’s essential attributes. 
By capturing critical details—such as objects, attributes, or contextual elements—these captions provide rich textual grounding information for the modification text generation.

Next, we supply the tuple $\bigl(x_i, c_{x_i}, y_i, c_{y_i}\bigr)$ to the MRA with a designed prompt $P_m$ to generate the modification text $t_i$:
\begin{equation}
t_i = \mathrm{MRA}\bigl(x_i, c_{x_i}, y_i, c_{y_i}, P_m\bigr).
\label{eq:text_generation_refined}
\end{equation}
By grounding the comparison in both the raw visual content and the semantic cues from captions, the MRA generates a more accurate and semantically coherent modifying text $t_i$. All the proposed prompts are provided in the Appendix \ref{prompt}.

\subsection{Fine-tuning VLM}
\label{section:finetune_vlm}
With the curated triplets \(\langle x_i, t_i, y_i \rangle\) in hand, we fine-tune a Vision-Language Model (VLM) to extract feature representations of compositional queries \(\langle x_i, t_i \rangle\) and align them with the features of their corresponding target images $y_i$. Figure~\ref{fig:framework} (b) provides an overview.

\paragraph{Composed Query and Target Image Features via Q-Former.}
We use Q-Former~\cite{li2023blip} to obtain features for both the composed queries and the target images. For the composed query \(\langle x_i, t_i \rangle\),  we first tokenize  \(t_i\) into text tokens and process \(x_i\) through a frozen image encoder to obtain image token features. These image and text token sequences, along with \(p\) learnable visual prompt tokens $\{{\boldsymbol{e}}^s\}_{s=1}^p$, are then fed into multiple Q-Former blocks, each incorporating self-attention and cross-attention mechanisms.
Lastly, the $p$ output query embeddings from the final Q-Former block, denoted as $\{\boldsymbol{u}_{i}^s\}_{s=1}^p$, serve as the final feature representation of the composed query.
These embeddings encode compositional information from both the reference image and its modifying text.

To encode the target image \(y_i\), we follow a similar procedure, except no text is included. Taking the image \(y_i\) and the learnable visual prompt tokens $\{{\boldsymbol{e}}^s\}_{s=1}^p$ as input, the output of the final Q-Former block, denoted as  $\{\boldsymbol{v}_{i}^s\}_{s=1}^p$, represents the feature representation for the target image \(y_i\).

\paragraph{Similarity Computation and InfoNCE Loss.}
We compute the similarity between the composed query \(\langle x_i, t_i \rangle\) and a candidate image \(x_j\) by first identifying the maximum cosine similarity at the token level, and then averaging:
\begin{equation}
\begin{aligned}
s_{ij} &= \frac{1}{p}\sum_{s=1}^p \max_{1 \le r \le p} \frac{(\boldsymbol{u}_{i}^s)^T \cdot {\boldsymbol{v}}_{j}^r}{\|\boldsymbol{u}_{i}^s\|_2 \|{\boldsymbol{v}}_{j}^r\|_2}.
\end{aligned}
\label{eq:similarity}
\end{equation}
Our objective is to make the similarity \(s_{ii}\) between the composed query \(\langle x_i, t_i \rangle\) and its corresponding target \(y_i\) larger than the similarity \(s_{ij}\) for any mismatched target \(y_j\), where \(j\neq i\). We thus adopt the  InfoNCE loss~\cite{infonce} where the similarity measure is  replaced by the maximum cosine similarity:
\begin{equation}
\mathcal{L} = \frac{1}{N} \sum_{i=1}^N \log{\frac{\exp(s_{ii}/\tau)}{\sum_{j=1}^N \exp(s_{ij}/{\tau})}},
\label{eq:infonce}
\end{equation}
where \(\tau\) is a learnable parameter, and \(N\) denotes the total number of triplets in the mini-batch.

By enforcing higher similarity scores for matching pairs through the proposed InfoNCE loss in Eq.~\eqref{eq:infonce}, the model learns to effectively capture fine-grained visual and textual cues necessary for aligning compositional queries with their corresponding target images. However, unlike the standard InfoNCE loss that typically operates on aggregate embeddings, our formulation employs a token-level maximum cosine similarity, introducing additional complexity whose theoretical implications remain unclear. To rigorously justify our design and ensure its optimization effectiveness and stable convergence, we next establish a theoretical connection between our proposed loss and the standard InfoNCE loss. 

\paragraph{Theoretical Analysis of Similarity Measures and InfoNCE Loss.}
Now, we show that our token-level maximum similarity formulation implicitly optimizes a lower bound of the standard InfoNCE objective, thus providing solid theoretical support for its practical effectiveness in training robust compositional image retrieval models.
First, we put forward a relatively strong hypothesis to facilitate a better theoretical characterization of the maximum cosine similarity. 
\newtheorem{assumption}{Assumption}[section]
\begin{assumption}
After a sufficient number of iterations, there exists a bijective $\sigma(\cdot) : (1,2,\cdots,p) \mapsto (1,2,\cdots,p) $ such that the following condition holds:
\begin{equation}
    \sigma(s) = {\rm \mathop{arg\ max}\limits_{r}} \frac{(\boldsymbol{u}_{i}^s)^T \cdot {\boldsymbol{v}}_{i}^r}{\|\boldsymbol{u}_{i}^s\|_2 \|{\boldsymbol{v}}_{i}^r\|_2}, \forall \ i.
\end{equation}\label{assump1}
\end{assumption}

Intuitively, Assumption~\ref{assump1} implies that, given sufficient training, the Q-Former is capable of optimally aligning each token-level embedding from the composed query with a unique corresponding embedding from its matched target image. This condition effectively characterizes an ideal scenario where the compositional alignment between query and target tokens is perfect. 
With this bijective $\sigma$, we define $\boldsymbol{U}_i = \frac{1}{\sqrt{p}} \left(\frac{( \boldsymbol{u}_{i}^1)^T}{\|\boldsymbol{u}_{i}^1\|_2}, \frac{(\boldsymbol{u}_{i}^2)^T}{\|\boldsymbol{u}_{i}^2\|_2}, \cdots, \frac{(\boldsymbol{u}_{i}^p)^T}{\|\boldsymbol{u}_{i}^p\|_2} \right)$,   
$\boldsymbol{V}_j = \frac{1}{\sqrt{p}}  \left((\frac{ (\boldsymbol{v}_{j}^{\sigma(1)})^T}{ \|{\boldsymbol{v}}_{j}^{\sigma(1)}\|_2}, \frac{ (\boldsymbol{v}_{j}^{\sigma(s)})^T}{ \|{\boldsymbol{v}}_{j}^{\sigma(s)}\|_2}, \cdots, \frac{ (\boldsymbol{v}_{j}^{\sigma(p)})^T}{ \|{\boldsymbol{v}}_{j}^{\sigma(p)}\|_2}\right)$. Further, we recover the standard infoNCE loss:
\begin{equation} \label{eq: standard infonce}
    \mathcal{L}^{s} = \frac{1}{N} \sum_{i=1}^N \log{\frac{\exp(\hat{s}_{ii}/\tau)}{\sum_{j=1}^N \exp(\hat{s}_{ij}/{\tau})}}, \ \hat{s}_{ij} = \boldsymbol{U}_{i}^{\top} \boldsymbol{V}_{j}.
\end{equation}
\begin{proposition}
With Assumption~\ref{assump1}, we can obtain that $\hat{s}_{ii} = s_{ii},\hat{s}_{ij} \leq s_{ij}$, and 
\begin{equation}
    \mathcal{L} = \frac{1}{N} \sum_{i=1}^N \log{\frac{\exp(s_{ii}/\tau)}{\sum_{j=1}^N \exp(s_{ij}/{\tau})}}  \leq \frac{1}{N} \sum_{i=1}^N \log{\frac{\exp(\hat{s}_{ii}/\tau)}{\sum_{j=1}^N \exp(\hat{s}_{ij}/{\tau})}} = \mathcal{L}^{s}.
\end{equation}
\end{proposition}

To estimate the gap between our loss and the infoNCE loss, we propose the following corollary:

\begin{corollary}  \label{corollary}
Suppose there exist constants $p_1$ and $p_2$ such that, after sufficient iterations, the ideal Q-Former satisfies  $s_{ii} \geq p_1$ and $s_{ij} \leq p_2$, i.e.,  matching similarities exceed a threshold while mismatches remain below another. We then have 
\begin{align}
    \mathcal{L}^{s}-\mathcal{L} \leq(N-1) \exp((p_2-p_1)/{\tau}).
\end{align}
\end{corollary}
From above insight, the InfoNCE loss with similarity measures utilized in our paper is essentially a lower bound of the standard InfoNCE loss. Consequently, the iterative optimization of our algorithm implicitly optimizes the standard InfoNCE objective. Prior studies~\cite{cho2024minibatch, koromilas2024bridging}, which analyzed optimal solutions of mini-batch InfoNCE loss, further indicate that our algorithm inherits similar optimality and convergence properties. Due to space limitation, we defer the detailed proof to the Appendix \ref{theoretical}.

\section{Experiments}

In this section, we present our experimental results to address the following research questions. 
\begin{itemize}[nosep]
    \item \textbf{RQ1}: How effective is the  proposed MRA-CIR method for the ZS-CIR task?
    \item \textbf{RQ2}: How does each component of MRA-CIR contribute to its performance?
    \item \textbf{RQ3}: How sensitive is MRA-CIR to the hyper-parameters? 
\end{itemize}

\subsection{Experimental Setting}
\subsubsection{Evaluation Dataset}
To comprehensively evaluate the performance of MRA-CIR across diverse CIR tasks, we used three public datasets: FashionIQ \cite{fashioniq}, CIRR \cite{cirr}, and CIRCO \cite{searle}. 
\textbf{FashionIQ:} This dataset contains fashion items across three categories: Dresses, Shirts, and Tops \& Tees. It features 36k validation triplets and is widely used for fashion-oriented CIR evaluations. Following prior studies \cite{pic2word,fti4cir}, we evaluated on the validation set due to the unavailability of the test set.
\textbf{CIRR:} CIRR consists of ~21k real-world images from NLVR2 \cite{NLVR2}, annotated to ensure that modifying texts are uniquely relevant to one target image pair. This eliminates false negatives, making CIRR a challenging benchmark for CIR models. Our evaluations used the test set with 4.1k triplets.
\textbf{CIRCO:} This dataset extends COCO \cite{COCO} to address false negatives by including multiple target images per sample. Each triplet comprises a reference image, modifying text, and multiple target images. We evaluated on its test set containing 800 samples, making it suitable for assessing multi-target retrieval.

\subsubsection{Implementation Details}
To ensure a fair comparison with prior approaches, we utilize 10k unlabeled image from the subset of ImageNet-1k~\cite{ImageNet} as the fine-tuning dataset. For our MRA, we employed \textbf{MiniCPM-VL-2\_6}~\cite{minicpm}.  
We use the BLIP2 model (ViT-L/14 version) \cite{li2023blip} as the base Vision-Language Model (VLM) for fine-tuning. The training process is conducted using the AdamW optimizer~\cite{AdamW} with an initial learning rate of \(\mathbf{1 \times 10^{-5}}\), which was reduced by a factor of 0.1 every 10 epochs.
We set the batch size to 128, and all experiments were implemented in PyTorch with fixed random seeds to ensure reproducibility. Furthermore, \( q_1 \) and \( q_2 \) were set at 51 and 60, respectively.
Moreover, the CIR performance is assessed in a zero-shot setting, where the fine-tuned VLM (trained on ImageNet-1k) is directly evaluated on three benchmark datasets without any further fine-tuning. All the experiments are executed on a single NVIDIA A100 GPU (40GB). Moreover, each experiment is repeated three times with different random seeds, and the reported results are averaged across these runs.

\vspace{-3pt}
\subsubsection{Evaluation Protocol}
\vspace{-3pt}
We employ standard evaluation protocols for each dataset, tailored to their unique characteristics. For the FashionIQ dataset, we used recall at rank $R@K, (k = 10, 50)$ as the evaluation metric. To gauge overall performance, we computed the average $R@K$ across all three categories.
For the CIRR dataset,  we use multiple metrics, including \(R@K\) (\(K = 1, 5, 10, 50\)), \(R_\text{subset}@K\) (\(K = 1, 2, 3\)), and the average of \(R@5\) and \(R_\text{subset}@1\). The subset metric evaluates the model's ability to identify semantically similar images while mitigating false negatives. For the CIRCO dataset, due to its multi-target nature, we adopt the mean Average Precision \(mAP@K\) (\(K = 5, 10, 25, 50\)) as the primary evaluation metric. This metric provides a fine-grained assessment of the model's ability to retrieve all relevant target images.

\begin{table*}[t]
  \centering 
  \vspace{-5pt}
  \caption{Performance comparison on FashionIQ dataset.}
  \resizebox{\linewidth}{!}{
  \begin{tabular}{c|l|cc|cc|cc|cc} 
  \toprule
  \multirow{2}{*}{Supervision} & \multirow{2}{*}{Methods} & \multicolumn{2}{c}{Shirt} & \multicolumn{2}{c}{Dress}& \multicolumn{2}{c}{TopTee} & \multicolumn{2}{c}{Average}\\
   \cmidrule(lr){3-4}
  \cmidrule(lr){5-6}
  \cmidrule(lr){7-8}
  \cmidrule(lr){9-10}
  & & R@10 & R@50& R@10 & R@50& R@10 & R@50& R@10 & R@50\\  \midrule
  \multirow{11}{*}{{\textsc{Zero-shot}}}
  &Image-only  &10.40&22.03&3.91&12.14&7.70&18.05&7.33&17.41\\
  &Text-only  &23.15&38.22&17.00&37.13&24.57&42.73&21.58&39.36\\
  &Image$+$Text& 20.90&37.88&11.25&28.50&18.40&35.29&16.85&33.89\\
  & Pic2Word& 26.20 & 43.60 & 20.00 & 40.20 & 27.90 & 47.40 & 24.70 & 43.70 \\
  & {iSEARLE-XL-OTI} & {31.80} & {50.20} & {24.19} & 45.12 & {31.72} & {53.29} & {29.24} & {49.54} \\
   & {iSEARLE-XL} & 28.75 & 47.84 & 22.51 & {46.36} & {31.31} & 52.68 & 27.52 & {48.96} \\ 
   & {FTI4CIR} & 31.35 &50.59 & 24.39 & 47.84 & 32.43 & 54.21 & 29.39 & 50.88 \\ 
   &Context-I2W  & 29.70 & 48.60  & 23.10 & 45.30  & 30.60 & 52.90  & 27.80 & 48.90  \\ 
   &CIReVL       & 29.49 & 47.40  & 24.79 & 44.76  & 31.36 & 53.65  & 28.55 & 48.57  \\ 
   &LinCIR       & 29.10 & 46.81  & 20.92 & 42.44  & 28.81 & 50.18  & 26.28 & 46.49  \\ 
    & MLLM-I2W & 27.3 & 46.5 & 29.9 & 48.6 & 33.8 & 55.2 & 30.3 & 50.1 \\
    & \cellcolor{cyan!10}\textbf{MRA-CIR}  &\cellcolor{cyan!10}\textbf{40.43}&\cellcolor{cyan!10}\textbf{60.20}& \cellcolor{cyan!10}\textbf{31.87}&\cellcolor{cyan!10}\textbf{54.23}&\cellcolor{cyan!10}\textbf{41.25}&\cellcolor{cyan!10}\textbf{62.51}&\cellcolor{cyan!10}\textbf{37.85} &\cellcolor{cyan!10}\textbf{58.98} \\ \midrule
\multicolumn{1}{c|}{CIRR}  &Combiner &23.7&39.4&17.2&37.9&24.1	&43.9&21.7&40.4 \\
\multicolumn{1}{c|}{Fashion-IQ}&Combiner &37.2&55.8&30.3&54.5&39.2&61.3&35.6&57.2\\ \bottomrule
  \end{tabular}
  }
  \label{tab:fashion}
\end{table*}

\begin{table*}[]
\centering 
\caption{Performance comparison on CIRR dataset.}
\resizebox{\linewidth}{!}{ 
\begin{tabular}{c|l|cccc|ccc|c}
\toprule
{Supervision} & {Method} & {R@1} & {R@5} & {R@10} & {R@50} & {R\textsubscript{subset}@1} & {R\textsubscript{subset}@2} & {R\textsubscript{subset}@3} & {Avg} \\ \midrule
\multirow{11}{*}{{ZERO-SHOT}} & Image-only &7.83& 24.51& 34.89&61.11& 20.99 & 41.30 & 60.84 &22.75 \\
 & Text-only & 20.31 & 43.98 &55.61&78.43&60.46 &80.87&90.92 &52.22\\
 & Image + Text & 10.55 & 32.53 & 45.47 &76.29 &29.93&53.86&72.48 &31.23\\
 & Pic2Word & 23.90 & 51.70 & 65.00 & 87.80 & - & - & - & - \\ 
 & iSEARLE-XL-OTI  &  {25.40} & {54.05} & {67.47} & {88.92} & - & - & - & - \\ 
 & iSEARLE-XL & 25.28 & 54.00 & 66.72 & 88.80 & - & - & - & - \\ 
 & FTI4CIR & {25.90} & {55.61} & {67.66} & {89.66} & {55.21} & {75.88} & {87.98} & {55.41} \\ 
 & CIReVL &24.55 &52.31 &64.92 &86.34 & 59.54 &79.88 &89.69 &- \\
 & MCL &26.22 &56.84 & 70.00 &91.35 &61.45 &81.61&91.93&59.15 \\ 
 & MLLM-I2W &28.3  &57.9  &70.2  &93.9  & - & - & - &-\\
 & \cellcolor{cyan!10} \textbf{MRA-CIR} &\cellcolor{cyan!10}\textbf{37.98}  & \cellcolor{cyan!10}\textbf{67.45}  &	\cellcolor{cyan!10}\textbf{78.07}  &	\cellcolor{cyan!10}\textbf{93.98} &\cellcolor{cyan!10}\textbf{64.17 } & \cellcolor{cyan!10}\textbf{83.01} &\cellcolor{cyan!10}\textbf{91.78}&\cellcolor{cyan!10}\textbf{65.81} \\ 
 \midrule
\multicolumn{1}{c|}{Fashion-IQ} & Combiner  & 21.11 & 50.96 & 64.75 & 87.95 & 48.63 & 71.90 & 86.24 & 49.80 \\
\multicolumn{1}{c|}{CIRR} & Combiner  & 31.61 & 62.22 & 75.23 & 93.52 & 60.63 & 80.84 & 90.99 & 61.42 \\ \bottomrule
\end{tabular}}
\label{tab:cirr}
\vspace{-13pt}
\end{table*}

\subsection{On Model Performance (RQ1)}
To evaluate the effectiveness of our proposed method, similar to previous methods \cite{mllmi2w,fti4cir}, we design three baseline variants using BLIP2 encoders: 1) {Image-only}: Encode the reference image and the candidate images using BLIP2’s visual encoder and then compute their feature similarity directly; 2) {Text-only}: Encode the modifying text and the candidate images through the BLIP2’s text and visual encoders and then measure similarity between them: 3) {Image + Text}: Average the features from the reference image and the modifying text into a single query representation, then compare it to the candidate image features. We also benchmark our approach against several state-of-the-art zero-shot CIR (ZS-CIR) methods to demonstrate generality and effectiveness, including Pic2Word~\cite{pic2word}, Context-I2W~\cite{contex12w}, LinCIR~\cite{lincir}, iSEARLE-XL~\cite{isearle}, FTI4CIR~\cite{fti4cir}, CIReVL~\cite{cirevl},  MLLM-I2W~\cite{mllmi2w}, and MCL~\cite{mcl}. Additionally, we also adopt the supervised method Combiner \cite{combiner} as baseline. We train this method on the popular CIR datasets FashionIQ (18K triplets) and CIRR (28K triplets), and then evaluate the resulting networks on all three target datasets.

\begin{wraptable}{r}{0.7\textwidth}
\vspace{-13pt}
\caption{Performance comparison on CIRCO dataset.}
\label{tab:ciro}
\resizebox{\linewidth}{!}{%
\begin{tabular}{c|l|cccc}
\toprule
{Supervision} & {Method} & {mAP@5} & {mAP@10} & {mAP@25} & {mAP@50} \\ \midrule
\multirow{12}{*}{{ZERO-SHOT}} 
& Image-only &2.59&3.2&3.98& 4.52 \\
& Text-only &3.36&3.79&4.4&4.76 \\
& Image + Text &6.67 &7.98 &9.69&10.56 \\
& Captioning & 8.33 & 8.98 & 10.17 & 10.75 \\
& Pic2Word  & 8.72 & 9.51 & 10.46 & 11.29 \\
& {iSEARLE-XL-OTI} & {11.31} & {12.67} & {14.46} & {15.34} \\
& {iSEARLE-XL} & {12.50} & {13.61} & {15.36} & {16.25} \\ 
& LinCIR & 12.59 &13.58 & 15.00 &15.85 \\
& FTI4CIR & {15.05} & {16.32} & {18.06} & {19.05} \\ 
& CIReVL & 18.57 & 19.01 & 20.89 & 21.80 \\
&MCL &17.67 &18.86 &20.80 & 21.68 \\
& \cellcolor{cyan!10}\textbf{MIR-CIR} &\cellcolor{cyan!10}\textbf{27.14} &  \cellcolor{cyan!10}\textbf{28.85}  & \cellcolor{cyan!10}\textbf{31.54}  &\cellcolor{cyan!10} \textbf{32.63} \\ 
\midrule
\multicolumn{1}{c|}{Fashion-IQ}
& Combiner  & 8.91 & 10.29 & 11.72 & 12.52 \\ 
\multicolumn{1}{c|}{CIRR} & Combiner  & 8.56 & 9.20 & 10.43 & 11.06 \\ 
\bottomrule
\end{tabular}
}
\vspace{-18pt}
\end{wraptable} 

Table~\ref{tab:fashion}, \ref{tab:cirr}, and \ref{tab:ciro} summarize the performance comparison across the three datasets: FashionIQ, CIRR, and CIRCO, respectively. Based on these results, we highlight the following key observations:

\paragraph{(1) Superiority over Zero-Shot Baselines.}
Our proposed \textbf{MRA-CIR} consistently surpasses all zero-shot baselines on all three datasets. For instance, on FashionIQ, MRA-CIR achieves an average improvement of $7.55\%$ in R@10 compared to the strongest baseline, MLLM-I2W. On CIRCO, it outperforms two methods that rely on an LLM at inference time, namely CIReVL and MCL, by $8.57\%$ and $9.47\%$ in R@10, respectively. These gains underscore our model’s more effective cross-modal alignment and compositional reasoning, validating its robustness under varying data conditions.
\paragraph{(2) Better than the Supervised Combiner.}
We also compare MRA-CIR against the Combiner network trained on different datasets (FashionIQ or CIRR) in a fully supervised manner. As shown in Table~\ref{tab:fashion}, when Combiner is trained on CIRR, our method yields a $16.15\%$ improvement in R@10. Even when Combiner is trained on FashionIQ, MRA-CIR maintains a notable margin of $2.25\%$. Additionally, Combiner exhibits weaker transfer performance across datasets; for example, a model trained on CIRR struggles considerably on FashionIQ. We attribute this to the domain-specific nature of its supervised triplets, which may lead to overfitting on particular label distributions or textual styles. In contrast, MRA-CIR—relying on automatically constructed triplets rather than manual annotations—demonstrates stronger domain adaptability, highlighting the broader generalization capabilities of our zero-shot approach.
\paragraph{(3) Efficacy of (M)LLM-Based Methods.}
Among the baselines, methods that incorporate Large Language Models (LLMs) or Multimodal LLMs (MLLMs) consistently rank higher overall. Their enhanced language understanding and reasoning abilities are beneficial for zero-shot composed retrieval tasks. The strong performance of these (M)LLM-based approaches aligns with our findings, as MRA-CIR also leverages a multimodal reasoning mechanism to capture nuanced relationships between image content and textual modifications. Taken together, these results reinforce the notion that powerful semantic understanding and reasoning are crucial for effective ZS-CIR solutions.

\subsection{On Ablation Study (RQ2)}
To validate the importance of each component in our MRA-CIR framework, we devise four ablated variants under the same training and inference protocols as the full model: (1) \textbf{Top-1 Target Selection (Top1).}
In this variant, for a reference image $x_i$, we select the target image $y_i$ that maximizes the similarity score with $x_i$.
(2) \textbf{Random Target Selection (RandTarget).} Instead of picking a moderately similar image, we randomly select a target image from the entire dataset.
(3) \textbf{w/o Caption.} This variant skips the captioning step and directly feeds $(x_i, y_i)$ into the MLLM to obtain the modification text. 
(4) \textbf{QwenMRA.} We replace the MiniCPM-based MLLM in MRA-CIR with the Qwen model, preserving all other components.

\begin{table*}[]
  \centering 
  \caption{Ablation study on the three datasets.}
  \label{tab:ablation}
  \resizebox{\linewidth}{!}{ 
  \begin{tabular}{l|cc|cc|cc|cc|cc} 
  \toprule
  \multirow{2}{*}{Methods} & \multicolumn{2}{c}{FashinIQ-Dress} & \multicolumn{2}{c}{FashinIQ-Shirt} & \multicolumn{2}{c|}{FashinIQ-TopTee}& \multicolumn{2}{c|}{CIRR} & \multicolumn{2}{c}{CIRO} \\
   \cmidrule(lr){2-3}
  \cmidrule(lr){4-5}
  \cmidrule(lr){6-7}
  \cmidrule(lr){8-9}
  \cmidrule(lr){10-11}
   & R@10 & R@50& R@10 & R@50& R@10 & R@50 & R@1 & R@5& mAP@5 & mAP@10\\  \midrule
Top1  &30.34&52.35&36.35&58.64&38.50&60.68&28.03& 58.76&16.17 &18.29\\
RandTarget  &24.83&47.94&30.91&48.08&33.70&56.24&35.33&62.31&5.24&5.58\\
w/o Caption& 31.82&53.29&38.17 & 59.81&40.74&62.51&35.85&66.63&25.40&27.18\\
QwenMRA &30.84&52.70&36.51&56.77&40.38&60.84&38.86&\textbf{69.57}&24.15&25.51\\
\cellcolor{cyan!10}\textbf{MRA-CIR} & \cellcolor{cyan!10}\textbf{31.87}&\cellcolor{cyan!10}\textbf{54.23}&\cellcolor{cyan!10}\textbf{40.43}&\cellcolor{cyan!10}\textbf{60.20}&\cellcolor{cyan!10}\textbf{41.25}&\cellcolor{cyan!10}\textbf{62.51}&\cellcolor{cyan!10}\textbf{37.79}&\cellcolor{cyan!10}68.67&\cellcolor{cyan!10}\textbf{25.77}&\cellcolor{cyan!10}\textbf{27.37}\\
\bottomrule
  \end{tabular}}
\end{table*}

We present the ablation experimental results in Table~\ref{tab:ablation} and highlight the observations as follows:
\paragraph{(1) Moderate Similarity Matters.}
Both Top1 and RandTarget perform worse than MRA-CIR, confirming the value of selecting \textit{moderately} similar image pairs. When the target is too similar (Top), the modification text provides only minor changes (e.g., subtle color variations), reducing the task to near image-to-image matching. Conversely, overly dissimilar pairs (RandTarget) push the retrieval process toward text-to-image matching, since the modifying text simply re-describes the target. In both extremes, the model tends to over-rely on a single modality, leading to biased representations and degraded performance. Hence, guiding the model with partially similar reference-target pairs fosters more robust compositional learning.

\paragraph{(2) Caption Guidance Enhances Text Quality.}
w/o Caption underperforms the full MRA-CIR, indicating the benefit of first generating captions before producing the modifying text. The intermediate captions $\boldsymbol{c}_i$ and $\boldsymbol{c}_i^t$ effectively highlight salient attributes for each image, enabling the MLLM to focus on relevant transformations. Consequently, this two-step approach yields higher-quality triplets $\langle x_i, \boldsymbol{t}_i, y_i\rangle$, thus improving final retrieval performance.

\paragraph{(3) Different MLLM Backbones Lead to Varying Domain Performance.}
We observe that MRA-CIR achieves stronger retrieval on FashionIQ and CIRCO, whereas QwenMRA excels on CIRR. One likely cause is that these MLLMs differ in their training data or architectural design, emphasizing different aspects of text generation and domain adaptation. Hence, each method shows strengths in certain datasets but lags behind in others.

\subsection{On Sensitivity of Hyper-Parameters (RQ3)}

\begin{figure*}[hbt]
    \vspace{-1em}
    \begin{minipage}[b]{0.32\linewidth}
      \centering
      \includegraphics[width=\linewidth]{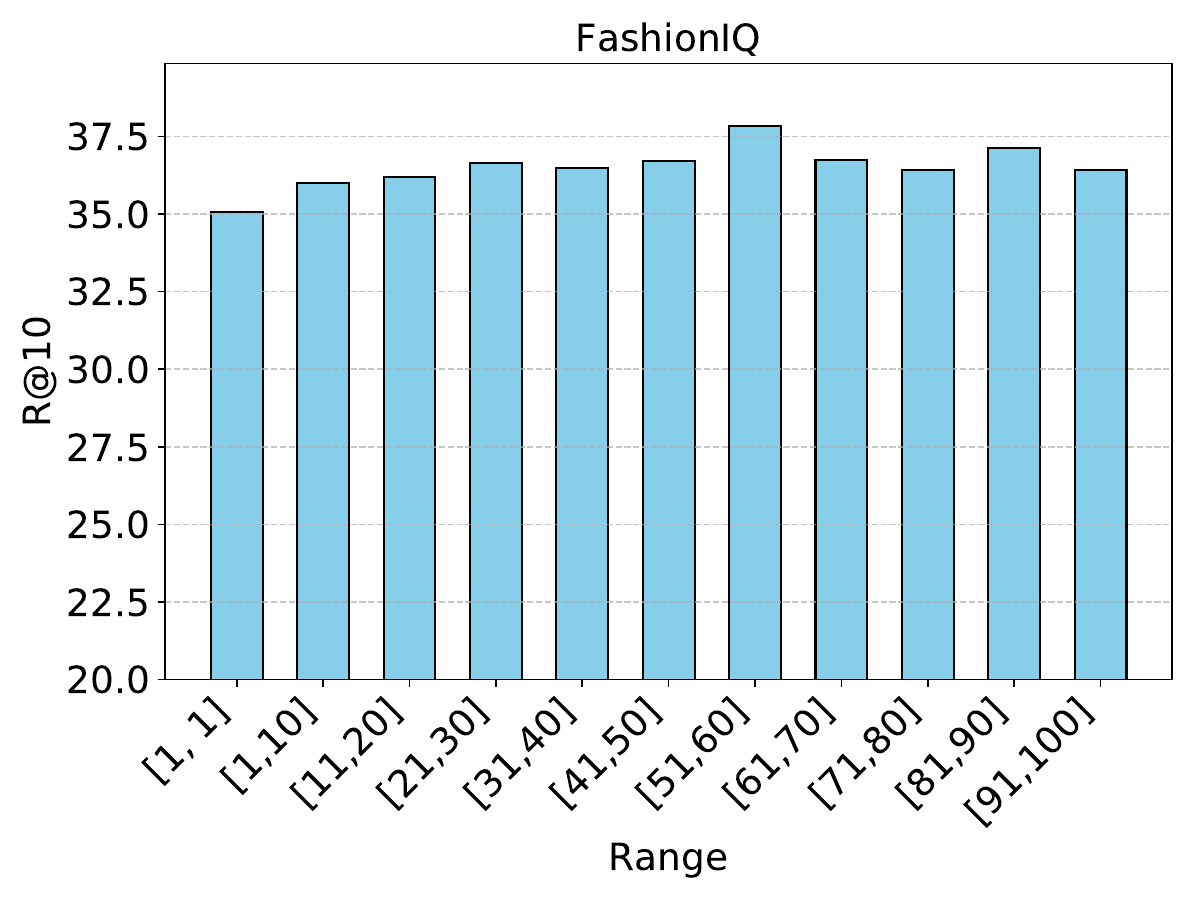}
    \end{minipage}
    \hfill
    \begin{minipage}[b]{0.32\linewidth}
      \centering
      \includegraphics[width=\linewidth]{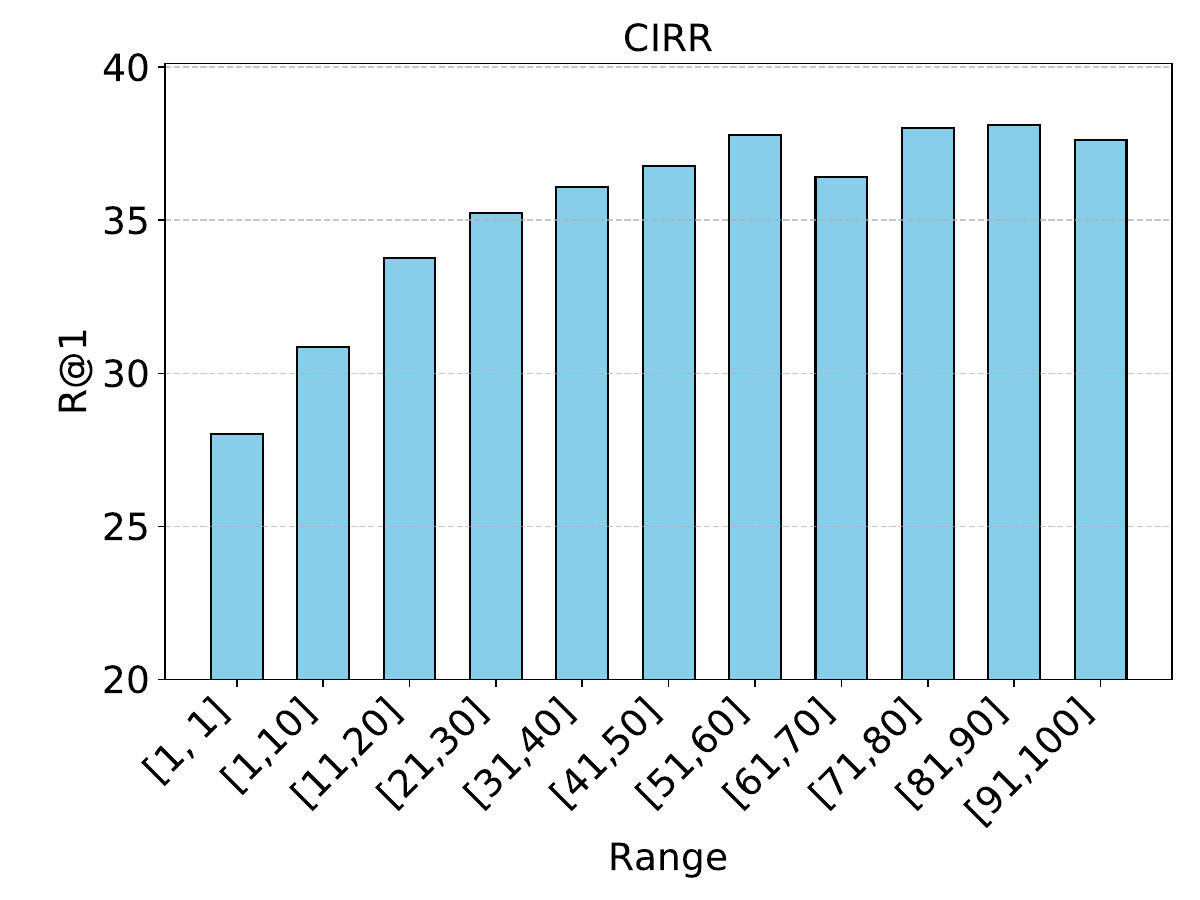}
    \end{minipage}
    \hfill
    \begin{minipage}[b]{0.32\linewidth}
      \centering
      \includegraphics[width=\linewidth]{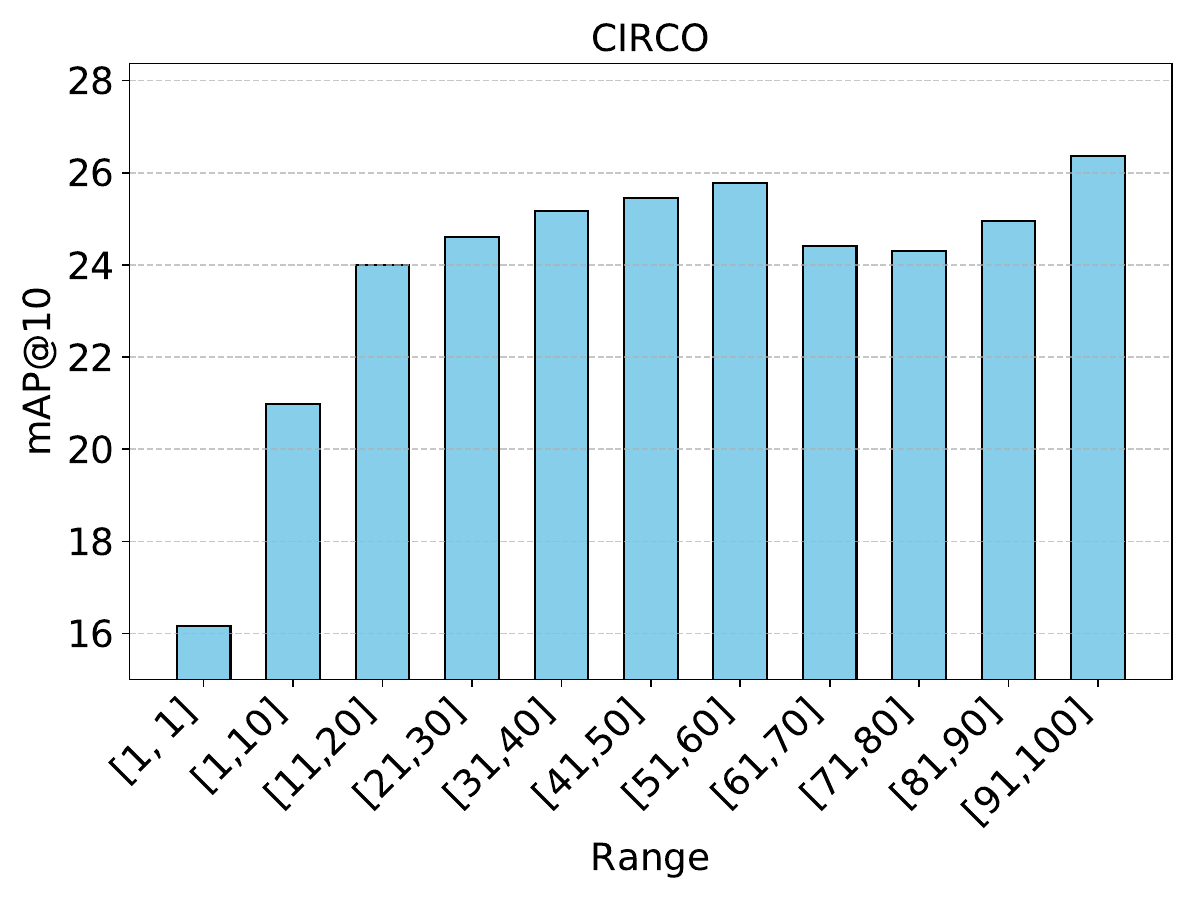}
    \end{minipage}
    \vspace{-0.5em}
    \captionof{figure}{
        Sensitivity analysis on the similarity ranking range $[q_1, q_2]$ for target image picking.
    }
    \label{fig:rank_range}
    \vspace{-10pt}
\end{figure*}

To evaluate the selection strategy for target images that ensures a moderate similarity with the reference image, we conduct experiments where we vary the similarity ranking range from which we pick the target image. The results are shown in Figure~\ref{fig:rank_range}.
When we always pick the top-ranked image (i.e., the most similar one) as the target, performance remains acceptable on FashionIQ but noticeably drops on CIRR and CIRO. 
Conversely, selecting targets with moderate similarity consistently yields better retrieval across these datasets. In particular, picking images ranked between the 51\textsuperscript{st} and 60\textsuperscript{th} most similar produces consistently great retrieval results across FashionIQ, CIRR, and CIRO. Therefore, we set $q_1=51$ and $q_2=60$ as 51 and 60 in other experiments.

\section{Conclusion}
In this work, we introduced MRA-CIR, a novel zero-shot composed image retrieval framework with a Multimodal Reasoning Agent. By directly constructing triplets 
<reference image, modification text, target image> from unlabeled images, our approach reduces the error propagation often encountered in existing methods that generate target text via large language models. Empirical evaluations on three benchmark datasets confirm that training on these automatically constructed triplets enables our model to more effectively capture the relationships between compositional queries and candidate images, thus outperforming the state-of-the-art baselines. 


\bibliographystyle{unsrtnat}
\bibliography{neurips_2025}
\newpage
\appendix



\section{Limitations}
\label{limitation}
Our framework relies on a Multimodal Language Model (MLLM) to generate the modifying text for each reference--target pair. Consequently, any misinterpretation of the image’s semantic content by the MLLM can lead to erroneous modification text, thereby propagating inaccuracies throughout the retrieval pipeline. Although our two-step approach—where we first produce captions and then generate the modification text—mitigates some of these errors, mistakes still occur in cases where the MLLM struggles with complex or ambiguous visual cues.

Another limitation stems from the unlabeled dataset itself, which may exhibit unbalanced distributions of reference--target differences (e.g., object additions/deletions, attribute changes, or background variations). Our current experiments do not explicitly address this imbalance, potentially causing the model to overfit certain transformation types while underrepresenting others. Future work could incorporate targeted data augmentation or sampling strategies to ensure a more uniform coverage of various transformation categories. Moreover, investigating more robust MLLMs or filtering mechanisms for text generation may further reduce the impact of incorrect semantic interpretations and enhance the overall retrieval performance.

\section{Broader Impact}
\label{broaderimpact}
This paper proposes a framework for zero-shot composed image retrieval (ZS-CIR) that leverages a multimodal large language model (MLLM) to construct supervision from unlabeled image pairs. By avoiding reliance on human-annotated triplets, the method can reduce annotation costs and increase accessibility for domains with limited labeled data. This may benefit practical applications such as visual search in e-commerce, content creation, and educational tools that rely on intuitive image-text interactions.

However, as the training supervision is generated automatically via an MLLM, the framework inherits any biases or errors present in the underlying language model. This may lead to semantically misleading or culturally biased retrieval behaviors, especially in ambiguous or underrepresented visual scenarios. In addition, the improved expressivity of retrieval systems raises concerns about potential misuse in surveillance or personal content retrieval without consent.

We mitigate these concerns by restricting our experiments to publicly available, non-sensitive datasets and disclosing model limitations. We encourage future deployment of such models to incorporate bias auditing, content filtering, and transparency mechanisms. The method is intended strictly for research use under responsible settings.

\begin{figure}[]
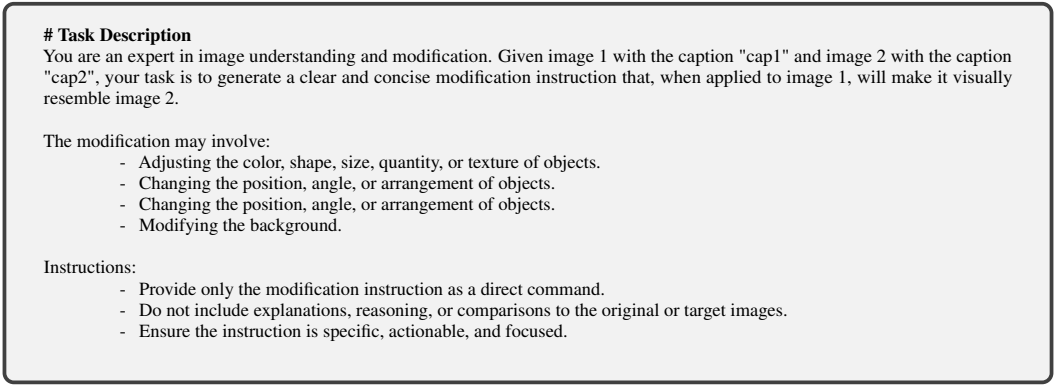

\scriptsize
\centering
\begin{tcolorbox}
\textbf{\# Task Description}\\
You are an expert in image understanding and modification. Given image 1 with the caption "{cap1}" and image 2 with the caption "{cap2}", your task is to generate a clear and concise modification instruction that, when applied to image 1, will make it visually resemble image 2. \\
\\
The modification may involve:
\begin{itemize}[nosep]
    \renewcommand{\labelitemi}{-}
    \item Adjusting the color, shape, size, quantity, or texture of objects.
    \item Changing the position, angle, or arrangement of objects.
    \item Changing the position, angle, or arrangement of objects.
    \item Modifying the background. \\
\end{itemize}

Instructions:
\begin{itemize}[nosep]
    \renewcommand{\labelitemi}{-}
    \item Provide only the modification instruction as a direct command.
    \item Do not include explanations, reasoning, or comparisons to the original or target images.
    \item Ensure the instruction is specific, actionable, and focused. \\
\end{itemize}
\end{tcolorbox}
\vspace{-0.1cm}
\caption{Caption based modification text generation Prompt template $P_m$.}
\vspace{-0.2cm}
\label{fig:mofication_generate}
\end{figure}

\begin{figure}[]
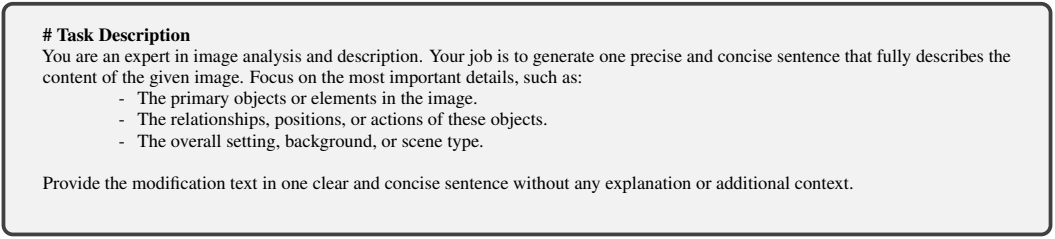

\scriptsize
\centering
\begin{tcolorbox}
\textbf{\# Task Description}\\
You are an expert in image analysis and description. Your job is to generate one precise and concise sentence that fully describes the content of the given image. Focus on the most important details, such as:
\begin{itemize}[nosep]
    \renewcommand{\labelitemi}{-}
    \item The primary objects or elements in the image.
    \item The relationships, positions, or actions of these objects.
    \item The overall setting, background, or scene type. \\
\end{itemize}

Provide the modification text in one clear and concise sentence without any explanation or additional context.\\
\end{tcolorbox}
\vspace{-0.1cm}
\caption{Image captioning Prompt template $P_c$.}
\vspace{-0.2cm}
\label{fig:caption}
\end{figure} 

\begin{figure}[]
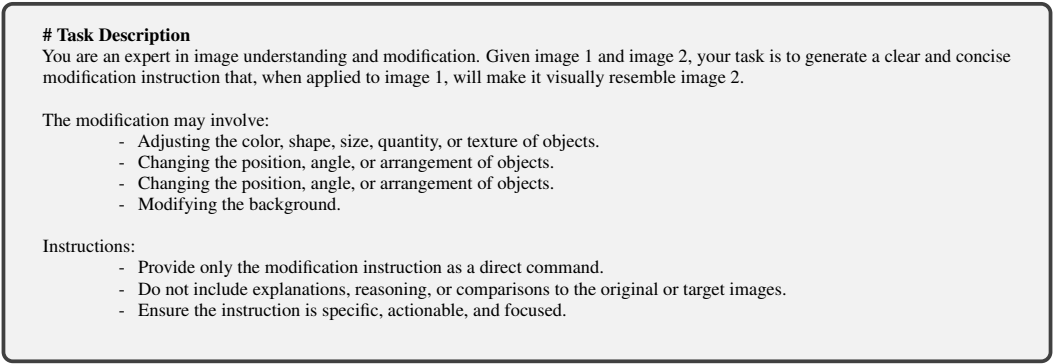

\scriptsize
\centering
\begin{tcolorbox}
\textbf{\# Task Description}\\
You are an expert in image understanding and modification. Given image 1 and image 2, your task is to generate a clear and concise modification instruction that, when applied to image 1, will make it visually resemble image 2.  \\
\\
The modification may involve:
\begin{itemize}[nosep]
    \renewcommand{\labelitemi}{-}
    \item Adjusting the color, shape, size, quantity, or texture of objects.
    \item Changing the position, angle, or arrangement of objects.
    \item Changing the position, angle, or arrangement of objects.
    \item Modifying the background. \\
\end{itemize}

Instructions:
\begin{itemize}[nosep]
    \renewcommand{\labelitemi}{-}
    \item Provide only the modification instruction as a direct command.
    \item Do not include explanations, reasoning, or comparisons to the original or target images.
    \item Ensure the instruction is specific, actionable, and focused. \\
\end{itemize}
\end{tcolorbox}
\vspace{-0.1cm}
\caption{Directly modification text generation Prompt template $P'_m$.}
\vspace{-0.2cm}
\label{fig:mofication_generate1}
\end{figure}

\section{Data Curation Prompt Templates}
\label{prompt}
In here, we illustrate the prompts for guiding the MRA to generate modification text and image caption in the Figure \ref{fig:mofication_generate} and  \ref{fig:caption}, respectively. Additionally, the corresponding prompts $\boldsymbol{P}'$ used for ablation study is shown in Figure \ref{fig:mofication_generate1}.

\section{Examples of Automatically Curated Triplets.}
\begin{figure*}[]
    \centering
    \includegraphics[width=0.9\linewidth]{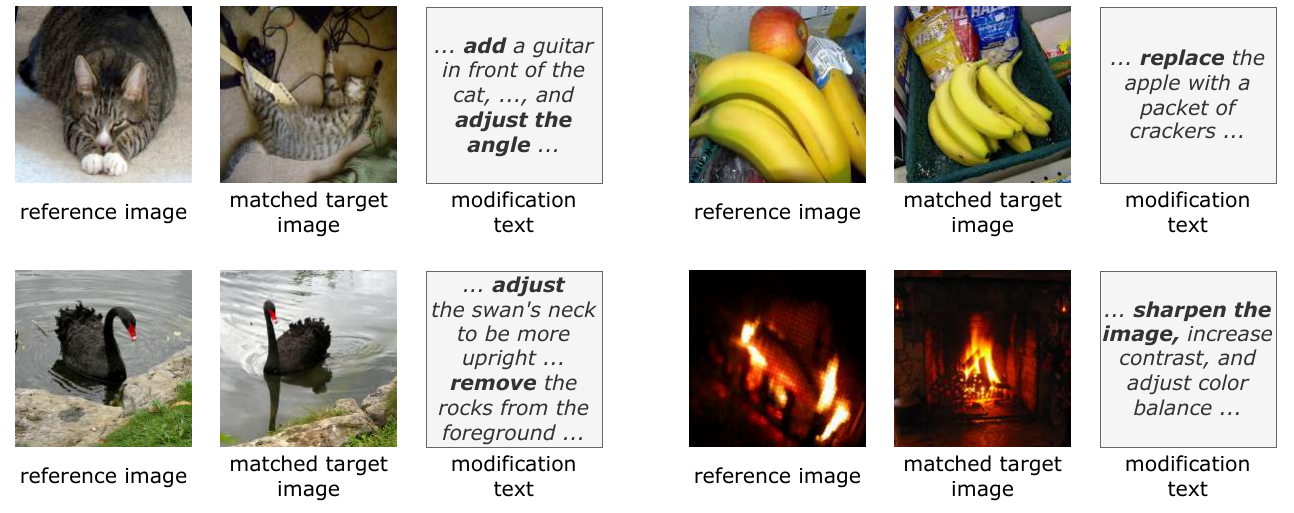}
    \vspace{-0.1cm}
    \caption{The examples of triplets curated by MRA based on unlabeled images.}
    \vspace{-0.2cm}
    \label{fig:mra_triplets}
\end{figure*}

Figure~\ref{fig:mra_triplets} illustrates some <reference image, modification text, target image> generated from unlabeled images through our MRA-based data construction pipeline. In each example, MRA identifies a target image that is neither trivially similar nor completely unrelated to the reference image, then produces a modification text describing the specific transformation required. These transformations range from adding or replacing key objects (e.g., a packet of crackers) to adjusting image attributes (e.g., angle, color balance, or background elements). Furthermore, these examples demonstrate that even with unlabeled images, MRA can pinpoint meaningful visual changes and express them in concise textual form, thus automatically curating high-quality triplets.

\section{Theoretical Analysis of the Loss Function}
\label{theoretical}
In our work, we use the formula ~(\ref{eq:similarity}) to calculate the similarity at the token level and then optimize the contrastive loss~(\ref{eq:infonce}). Compared with the usual cosine similarity, the maximum cosine similarity  can better describe the fine-grained semantic information in the embedded representation. However, the complex expression makes it very difficult to analyze the theoretical mechanism. In order to have an intuitive understanding of the nature of this similarity measurement, we propose Assumption~\ref{assump1} to conduct our analysis.

This assumption requires that Q-Former, after sufficient training, has a strong feature extraction and matching ability for the composed query and the corresponding image. For a composed query embedding, there exists a unique and definite image embedding corresponding to it. Both encode the same feature and have the highest similarity. With Assumption~\ref{assump1}, We can obtain the property of the maximum cosine similarity:
\begin{align}
    s_{ij} &= \frac{1}{p}\sum_{s=1}^p \max_{r} \frac{(\boldsymbol{u}_{i}^s)^T \cdot {\boldsymbol{v}}_{j}^r}{\|\boldsymbol{u}_{i}^s\|_2 \|{\boldsymbol{v}}_{j}^r\|_2} = \frac{1}{p}\sum_{s=1}^p  \frac{(\boldsymbol{u}_{i}^s)^T \cdot {\boldsymbol{v}}_{i}^{\sigma(s)}}{\|\boldsymbol{u}_{i}^s\|_2 \|{\boldsymbol{v}}_{i}^{\sigma(s)}\|_2} := \hat{s}_{ii}, \notag\\ 
    s_{ij} &= \frac{1}{p}\sum_{s=1}^p \max_{r} \frac{(\boldsymbol{u}_{i}^s)^T \cdot {\boldsymbol{v}}_{j}^r}{\|\boldsymbol{u}_{i}^s\|_2 \|{\boldsymbol{v}}_{j}^r\|_2} \leq \frac{1}{p}\sum_{s=1}^p  \frac{(\boldsymbol{u}_{i}^s)^T \cdot {\boldsymbol{v}}_{j}^{\sigma(s)}}{\|\boldsymbol{u}_{i}^s\|_2 \|{\boldsymbol{v}}_{j}^{\sigma(s)}\|_2} := \hat{s}_{ij} .
\end{align}
Then we have:
\begin{equation}
    \frac{1}{N} \sum_{i=1}^N \log{\frac{\exp(s_{ii}/\tau)}{\sum_{j=1}^N \exp(s_{ij}/{\tau})}} = \frac{1}{N} \sum_{i=1}^N \log{\frac{\exp(\hat{s}_{ii}/\tau)}{\sum_{j=1}^N \exp(s_{ij}/{\tau})}} \leq \frac{1}{N} \sum_{i=1}^N \log{\frac{\exp(\hat{s}_{ii}/\tau)}{\sum_{j=1}^N \exp(\hat{s}_{ij}/{\tau})}}.
\end{equation}
By arranging the embedding representations, i.e., $\boldsymbol{U}_i = \frac{1}{\sqrt{p}} \left(\frac{( \boldsymbol{u}_{i}^1)^T}{\|\boldsymbol{u}_{i}^1\|_2}, \frac{(\boldsymbol{u}_{i}^2)^T}{\|\boldsymbol{u}_{i}^2\|_2}, \cdots, \frac{(\boldsymbol{u}_{i}^p)^T}{\|\boldsymbol{u}_{i}^p\|_2} \right)$,  
$\boldsymbol{V}_j = \frac{1}{\sqrt{p}}  \left((\frac{ (\boldsymbol{v}_{j}^{\sigma(1)})^T}{ \|{\boldsymbol{v}}_{j}^{\sigma(1)}\|_2}, \frac{ (\boldsymbol{v}_{j}^{\sigma(s)})^T}{ \|{\boldsymbol{v}}_{j}^{\sigma(s)}\|_2}, \cdots, \frac{ (\boldsymbol{v}_{j}^{\sigma(p)})^T}{ \|{\boldsymbol{v}}_{j}^{\sigma(p)}\|_2}\right)$, we recover the standard infoNCE loss:
\begin{equation} 
    \mathcal{L}^{s}=\frac{1}{N}\sum_{i = 1}^{N}\log\frac{\exp(\boldsymbol{U}_{i}^{\top}\boldsymbol{V}_{i}/{\tau})}{\sum_{j = 1}^{N} \exp(\boldsymbol{U}_{i}^{\top} \boldsymbol{V}_{j}/{\tau})} = \frac{1}{N} \sum_{i=1}^N \log{\frac{\exp(\hat{s}_{ii}/\tau)}{\sum_{j=1}^N \exp(\hat{s}_{ij}/{\tau})}}.
\end{equation}
Through the above analysis, we can obtain that the optimization objective~\ref{eq:infonce} actually gives a lower bound of the standard infoNCE loss ~\ref{eq: standard infonce}. To accurately estimate the gap between the optimization objective and the standard infoNCE loss, we will propose another assumption. In the actual training process, the optimization algorithm increases the matching similarity while reducing the mismatch similarity. We hope that the ideal Q-Former satisfies that the similarity between matching samples will be greater than a threshold, and the similarity between mismatch samples will be less than another threshold. Then we can obtain the proof of Corollary~\ref{corollary}
\begin{align}
    \mathcal{L}^{s}-\mathcal{L} &=  \frac{1}{N} \sum_{i=1}^N \log{\frac{\sum_{j=1}^N \exp({s}_{ij}/{\tau})}{\sum_{j=1}^N \exp(\hat{s}_{ij}/{\tau})}} \notag\\
    &=  \frac{1}{N} \sum_{i=1}^N \log{\frac{(\sum_{j=1}^N \exp({s}_{ij}/{\tau})-\sum_{j=1}^N \exp(\hat{s}_{ij}/{\tau}))+\sum_{j=1}^N \exp(\hat{s}_{ij}/{\tau})}{\sum_{j=1}^N \exp(\hat{s}_{ij}/{\tau})}} \notag\\
    &\leq \frac{1}{N} \sum_{i=1}^N \frac{\sum_{j=1}^N \exp({s}_{ij}/{\tau})-\sum_{j=1}^N \exp(\hat{s}_{ij}/{\tau})}{\sum_{j=1}^N \exp(\hat{s}_{ij}/{\tau})} \notag\\
    &= \frac{1}{N} \sum_{i=1}^N \frac{\sum_{j=1,j\neq i}^N \exp({s}_{ij}/{\tau})-\sum_{j=1,j\neq i}^N \exp(\hat{s}_{ij}/{\tau})}{\sum_{j=1}^N \exp(\hat{s}_{ij}/{\tau})} \notag\\
    &\leq(N-1) \exp((p_2-p_1)/{\tau})
\end{align}
Therefore, the iterative process of our algorithm can be regarded as an implicit optimization of the standard infoNCE loss.

Suppose there are a total of M images in the training set, the dimension of embedding vectors is d, and the infoNCE loss of all $\binom{N}{B}$ minibatches has been optimized, the previous works \citep{cho2024minibatch, koromilas2024bridging} provided the following lemma for the global optimal solution of the $\tbinom{M}{N}$ mini-batch infoNCE objective:

\newtheorem{lemma}{Lemma}
\begin{lemma}
Suppose $N\geq2 , \|\boldsymbol{U}_{i}\|_2 = \|\boldsymbol{V}_{i}\|_2 = 1.$ When $d\cdot p\geq M - 1$,   the solutions $(U,V)$ for the $\binom{N}{B}$ mini - batch optimization problem satisfies the following: 

(i) $\{\boldsymbol{U}_{i}\}_{i = 1}^{M}$ forms a simplex equiangular tight frame (ETF) , i.e., $\boldsymbol{U}_{i}^T\boldsymbol{U}_{j} = -\frac{1}{M-1}, \forall i\neq j$ 

(ii) $\boldsymbol{U}_{i}=\boldsymbol{V}_{i}$ for all $i\in[M]$.
\end{lemma}

Specifically, when $d\cdot p\geq M - 1$, the global maximum of $\mathcal{L}^{s}$ is achieved when $\{\boldsymbol{U}_{i}\}_{i = 1}^{M}$ form a ETF and $\boldsymbol{U}_{i}=\boldsymbol{V}_{i}$ for all $i = 1,\cdots,M$. This means that the feature vectors arrange themselves in a highly structured way, which is the essence of the Neural Collapse phenomenon \citep{LU2022224}. The intuitive explanation for this phenomenon is that the embedding of the matching image are exactly the same as the query embedding, while all the unmatched image embeddings are uniformly far away from the query embedding.

\begin{figure*}[]
    \centering
    \includegraphics[width=0.9\linewidth]{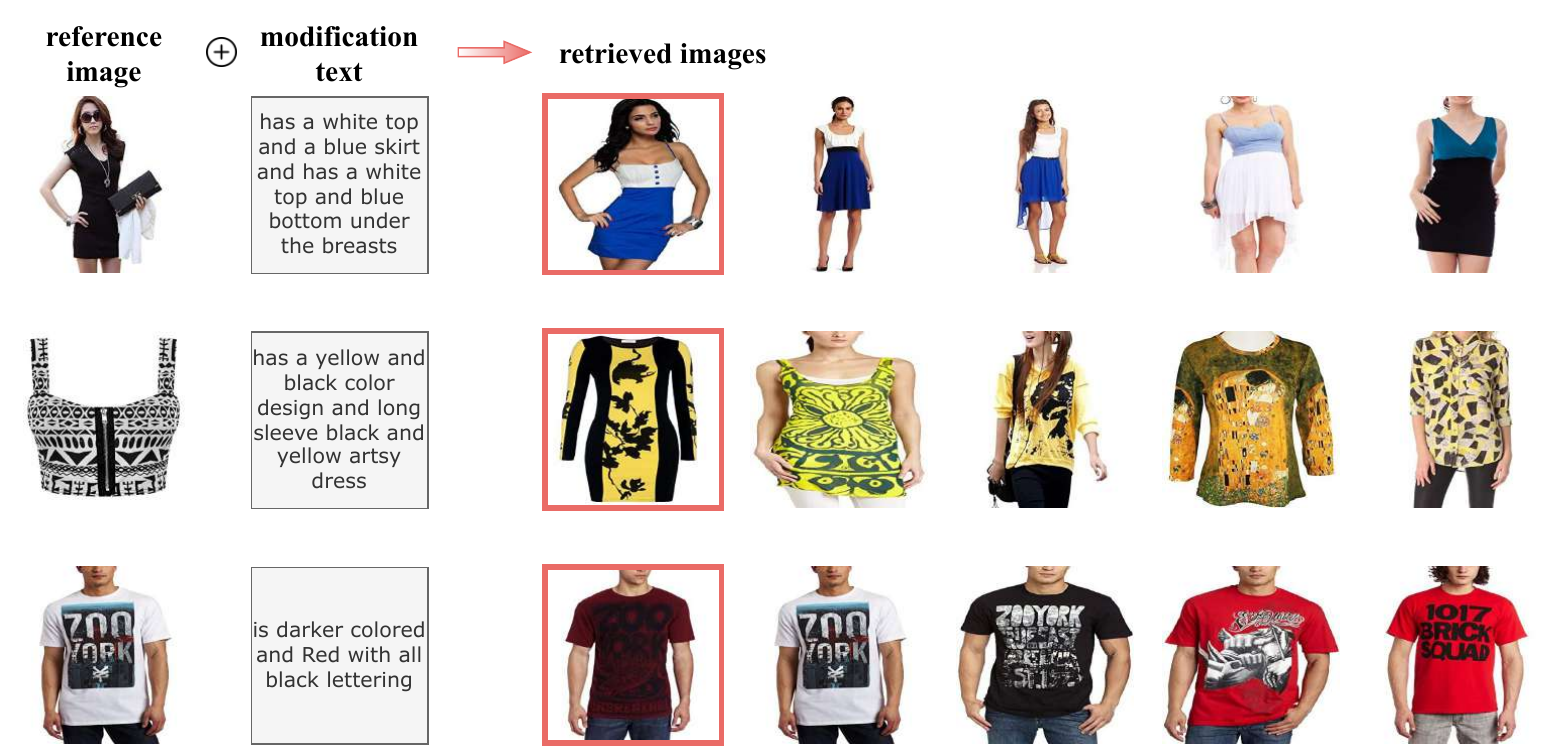}
    \caption{Retrieved results on the FashionIQ dataset. The target image is marked with the red box.}
    \label{fig:case_fashion}
\end{figure*}

\begin{figure*}[]
    \centering
    \includegraphics[width=0.9\linewidth]{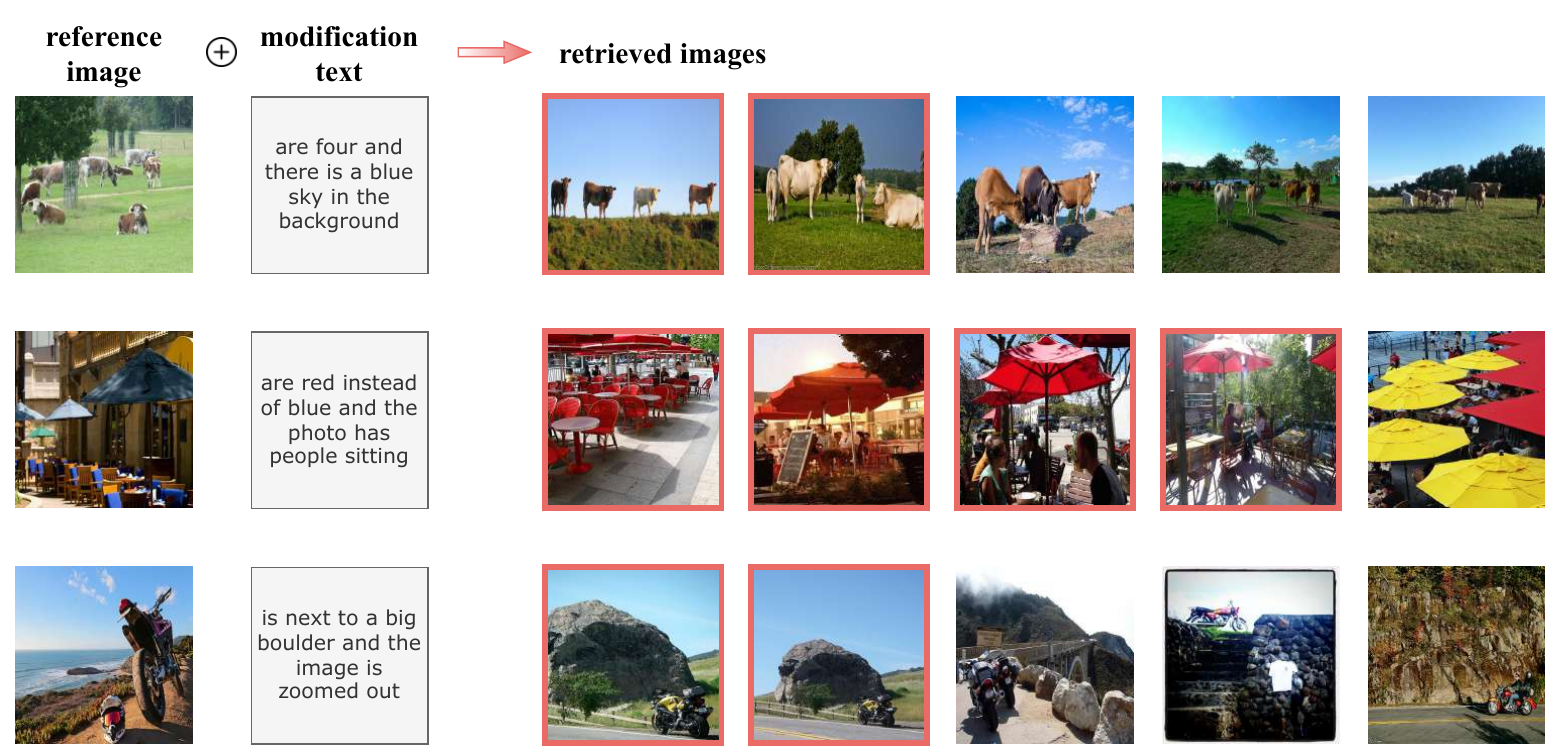}
    \caption{Retrieved results on the CIRCO dataset. The target images are marked with the red box.}
    \label{fig:case_circo}
\end{figure*}

\begin{figure*}[]
    \centering
    \includegraphics[width=0.9\linewidth]{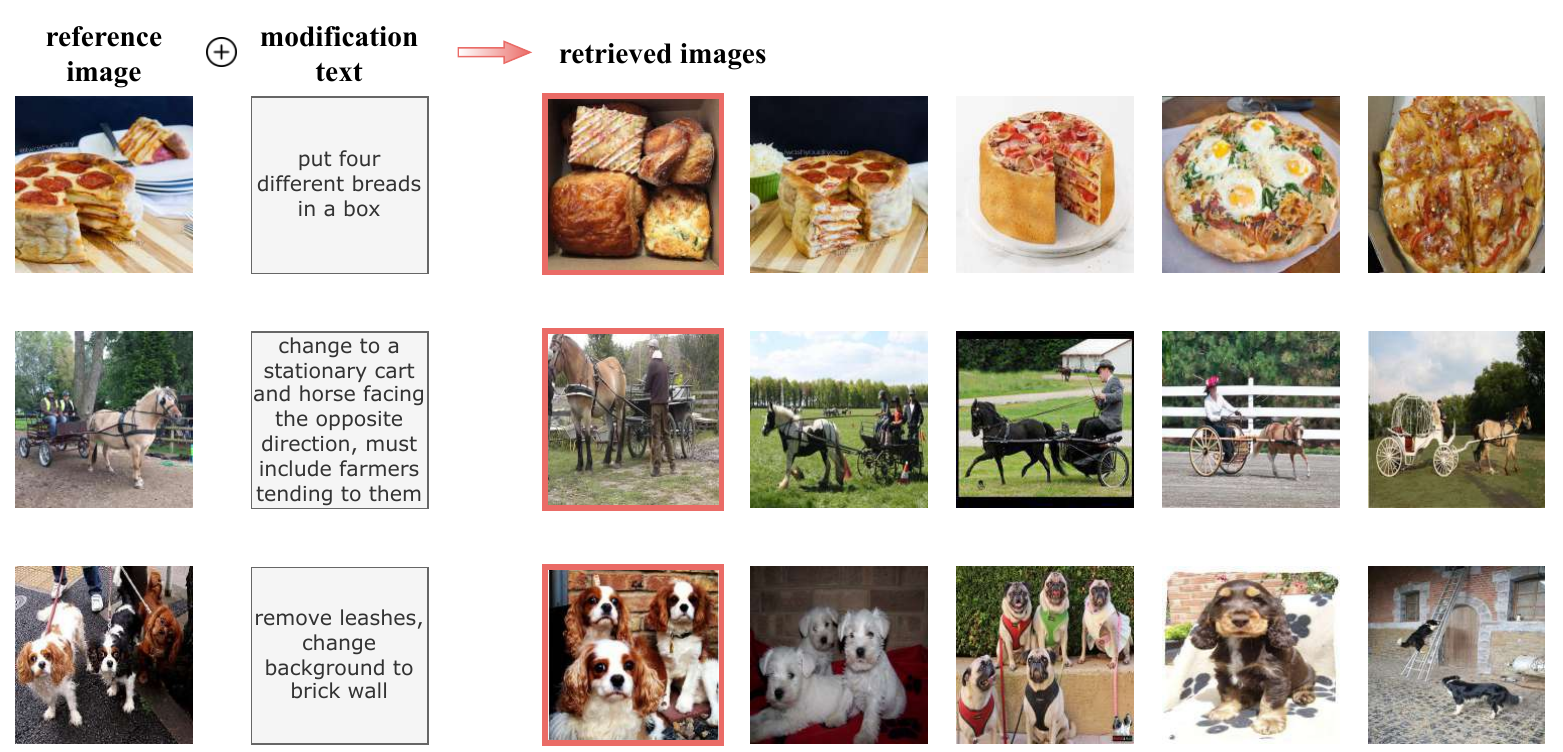}
    \caption{Retrieved results on the CIRR dataset. The target image is marked with the red box.}
    \label{fig:case_cirr}
\end{figure*}
\section{Case Study}
In Figures~\ref{fig:case_fashion}, \ref{fig:case_circo}, and \ref{fig:case_cirr}, we present qualitative retrieval examples on the FashionIQ, CIRCO, and CIRR datasets, respectively. Each figure illustrates one reference image, the associated modification text, and our top retrieved images, with the correct target(s) outlined in red.

On the \textit{FashionIQ} dataset (Figure~\ref{fig:case_fashion}), our approach accurately ranks the correct target at the top, even when the modifications involve intricate attributes such as color schemes or pattern details. This outcome indicates that our method effectively captures the fine-grained compositional cues needed for precise retrieval in the fashion domain.

For the \textit{CIRCO} dataset (Figure~\ref{fig:case_circo}), where each query may match multiple valid target images, our model successfully locates the correct targets in the top few positions. Despite the increased complexity arising from broader visual diversity, the retrieved images demonstrate that our compositional reasoning mechanism remains robust, accommodating various target appearances that align with the query instructions.

Finally, on the \textit{CIRR} dataset (Figure~\ref{fig:case_cirr}), our method again highlights the correct target images within the top ranks. These cases often feature more abstract semantic shifts—such as modifying scene context or adding specific attributes—yet the model consistently interprets the textual modifications and reference images to identify the intended targets.

Overall, these qualitative results confirm that our method excels at handling a wide range of compositional modifications, from subtle fashion details to context-rich scene variations, thus underscoring its strong generalization across different domains in zero-shot composed image retrieval.

\end{document}